%% file: aaai21.tex
\pdfoutput=1
\def\year{2021}\relax
\documentclass[letterpaper]{article} 
\usepackage{aaai21}  
\usepackage{times}  
\usepackage{helvet} 
\usepackage{courier}  
\usepackage[hyphens]{url}  
\usepackage{graphicx} 
\usepackage{natbib}  
\usepackage{caption} 
\usepackage{subcaption}
\usepackage[inline]{enumitem}
\usepackage{multirow}
\usepackage[switch]{lineno}  
\setlist[itemize]{noitemsep, topsep=0pt}
\title{SCAN: A Spatial Context Attentive Network for Joint Multi-Agent Intent Prediction}
\author{

    Jasmine Sekhon\textsuperscript{\rm 1}, Cody Fleming\textsuperscript{\rm 1, 2}
    
}
\affiliations{
\textsuperscript{\rm 1} University of Virginia, \\ \textsuperscript{\rm 2} Iowa State University \\
js3cn@virginia.edu, flemingc@iastate.edu

}

\begin{document}
\maketitle

\input{abstract}
\section{Introduction}\label{introduction}
\input{introduction}

\section{Related Work}\label{related_work}
\input{related_work}

\section{Proposed Approach}\label{approach}
\input{approach}
\section{Experimental Evaluation} 
\input{experiments}

\section{Conclusion and Future Work}
\input{conclusion}

\section*{Ethical Impact}
\input{ethicalimpact}
\section*{Acknowledgements}
This material is based upon work supported in part by the National Science Foundation under Grant No. CNS: 1650512, conducted in the NSF IUCRC Center of Visual and Decision Informatics, through the guidance of Leidos Corporation.

\bibliography{aaai21.bib}

\end{document}

%% file: abstract.tex
\begin{abstract}
    Safe navigation of autonomous agents in human centric environments requires the ability to understand and predict motion of neighboring pedestrians. However, predicting pedestrian intent is a complex problem. Pedestrian motion is governed by complex social navigation norms, is dependent on neighbors' trajectories, and is multimodal in nature. In this work, we propose \textbf{SCAN}, a \textbf{S}patial \textbf{C}ontext \textbf{A}ttentive \textbf{N}etwork that can jointly predict socially-acceptable multiple future trajectories for all pedestrians in a scene. SCAN encodes the influence of spatially close neighbors using a novel spatial attention mechanism in a manner that relies on fewer assumptions, is parameter efficient, and is more interpretable compared to state-of-the-art spatial attention approaches. Through experiments on several datasets we demonstrate that our approach can also quantitatively outperform state of the art trajectory prediction methods in terms of accuracy of predicted intent. 
\end{abstract}

%% file: introduction.tex
Modes of autonomous navigation are increasingly being adopted in land, marine and airborne vehicles. In all these domains, these autonomous vehicles are often expected to operate in human-centric environments (e.g. social robots, self-driving cars, etc.). When humans are navigating in crowded environments, they follow certain implicit rules of social interaction. As an example, when navigating in crowded spaces like sidewalks, airports, train stations, and others, pedestrians attempt to navigate safely while avoiding collision with other pedestrians, respecting others' personal space, yielding right-of-way, etc. Any autonomous agent attempting to navigate safely in such shared environments must be able to model these social navigation norms and understand neighbors' motion as a function of such complex spatial interactions. In this work, we aim to understand pedestrian interactions and model these towards \emph{jointly} predicting future trajectories for multiple pedestrians navigating in a scene. The contributions of our work are three-fold:
\begin{itemize}[noitemsep]
    \item We introduce a novel spatial attention mechanism to model spatial influence of neighboring pedestrians in a manner that relies on fewer assumptions, is parameter efficient, and interpretable. We encode the spatial influences experienced by a pedestrian at a point of time into a \emph{spatial context} vector. 
    \item We propose \textbf{SCAN}, a \textbf{S}patial \textbf{C}ontext \textbf{A}ttentive \textbf{N}etwork, that jointly predicts trajectories for all pedestrians in the scene for a future time window by attending to spatial contexts experienced by them individually over an observed time window. 
    \item Since human motion is multimodal, we extend our proposed framework to predicting multiple socially feasible paths for all pedestrians in the scene. 
\end{itemize}

%% file: related_work.tex
 Since a key contribution of our work is the ability of our proposed framework to model spatial interactions between neighboring pedestrians in a novel manner, we briefly discuss how existing trajectory forecasting methods encode spatial influences while predicting pedestrian intent. 

Traditional methods have relied on hand-crafted functions and features to model spatial interactions. For instance, the Social Forces model~\cite{Helbing_1995} models pedestrian behavior with attractive forces encouraging moving towards their goal and repulsive forces discouraging collision with other pedestrians. Similarly, ~\cite{6909680} and ~\cite{7298971} proposed trajectory forecasting approaches that rely on features extracted from human trajectories or human attributes. Such methods are limited by the need to hand craft features and attributes and their simplistic models and lack generalizability to complex crowded settings. Further, they only model immediate collision-avoidance behavior and do not consider interactions that may occur in the more distant future. 

More recently, deep learning based frameworks are being used to model spatial interactions between pedestrians. 
LSTM-based (Long short-term memory) approaches are well-suited to predict pedestrian trajectories owing to the sequential nature of the data. Consequently, several LSTM-based approaches have been proposed and successfully applied to predict pedestrian intent in the past. Alahi \textit{et. al.} proposed Social LSTM~\cite{Alahi_2016_CVPR} that uses a social pooling layer to encode spatial influences from neighboring pedestrians within an assumed spatial grid. More recently, Gupta \textit{et. al.} proposed Social GAN~\cite{gupta2018social}, which goes beyond modeling only local interactions within a fixed spatial grid, and considers influence of every other pedestrian in the scene on the pedestrian of interest. However, they use maxpooling, which causes all neighboring agents to have an identical representation towards predicting intent for a pedestrian of interest. Therefore, their method treats the influence of all agents on each other uniformly. SophieGAN~\cite{Sadeghian_2019_CVPR} eliminates this problem by using a sorting mechanism based on distance to create a feature representation to encode spatial influences of neighbors. This causes each neighbor to have its unique feature representation, and hence, all neighbors have different spatial influences on a pedestrian. However, two neighbors at the same distance from a pedestrian may have different spatial influences. For instance, a neighbor at a certain distance from the pedestrian of interest, but not in line-of-sight, may have negligible influence on it, in comparison to another neighbor at the same distance but approaching it head-on. Such factors, like orientation, are therefore, imperative towards encoding spatial influence. 

Graph Attention Networks, proposed by Velickovic \textit{et. al.}~\cite{Velickovic2018GraphAN}, allow for application of self-attention over any type of structured data that can be represented as a graph. Pedestrian interactions can be naturally represented as graphs, where nodes are pedestrians and edges are spatial interactions. Several attention-based graph approaches~\cite{Kosaraju2019SocialBiGATMT, Mohamed_2020_CVPR, amirian2019social,Vemula2018SocialAM} are used for modeling spatial interactions. At a very high level, graph attention networks compute weights for edges by using scoring mechanisms (e.g. dot product of the hidden states of the nodes connected by the edge). Such a scoring mechanism does not consider the effect of features such as distances, relative orientations, etc. on the spatial influence of a neighbor. In ~\cite{Vemula2018SocialAM}, Vemula \textit{et. al.} proposed Social Attention that takes into account the effect of this relative orientation towards spatial influence by encoding this information in spatial edges of a spatio-temporal graph. Similarly, Social Ways~\cite{amirian2019social} computes spatial influence of a neighbor as the scalar product of the hidden state of the neighbor and a feature vector that contains orientation features. A key disadvantage of such approaches is that the number of trainable parameters towards computing spatial interactions are proportional to the number of nodes in the graph. As we explain later, our proposed spatial interaction mechanism is able to model spatial influence such that the number of trainable parameters are independent of the number of nodes/pedestrians in the graph. Our proposed approach models spatial influence in a manner that is parameter efficient and more interpretable compared to existing approaches.

%% file: approach.tex
Given $N$ pedestrians present in a given frame at the start of an observation time window, from $t_{0}$ to $T_{obs}$, our goal is to \emph{jointly predict} socially plausible trajectories for each of the $N$ pedestrians in the scene over a time window in the future, from $T_{obs}+1$ to $T_{pred}$. The trajectory of a pedestrian $p$ at time $t$ is denoted by ($x^{p}_{t}$, $y^{p}_{t}$). \\
\input{model_architecture}
\input{spatial_attention_mechanism}
\input{multimodalbehavior}


%% file: model_architecture.tex
\begin{figure}
\centering
\includegraphics[width=\linewidth]{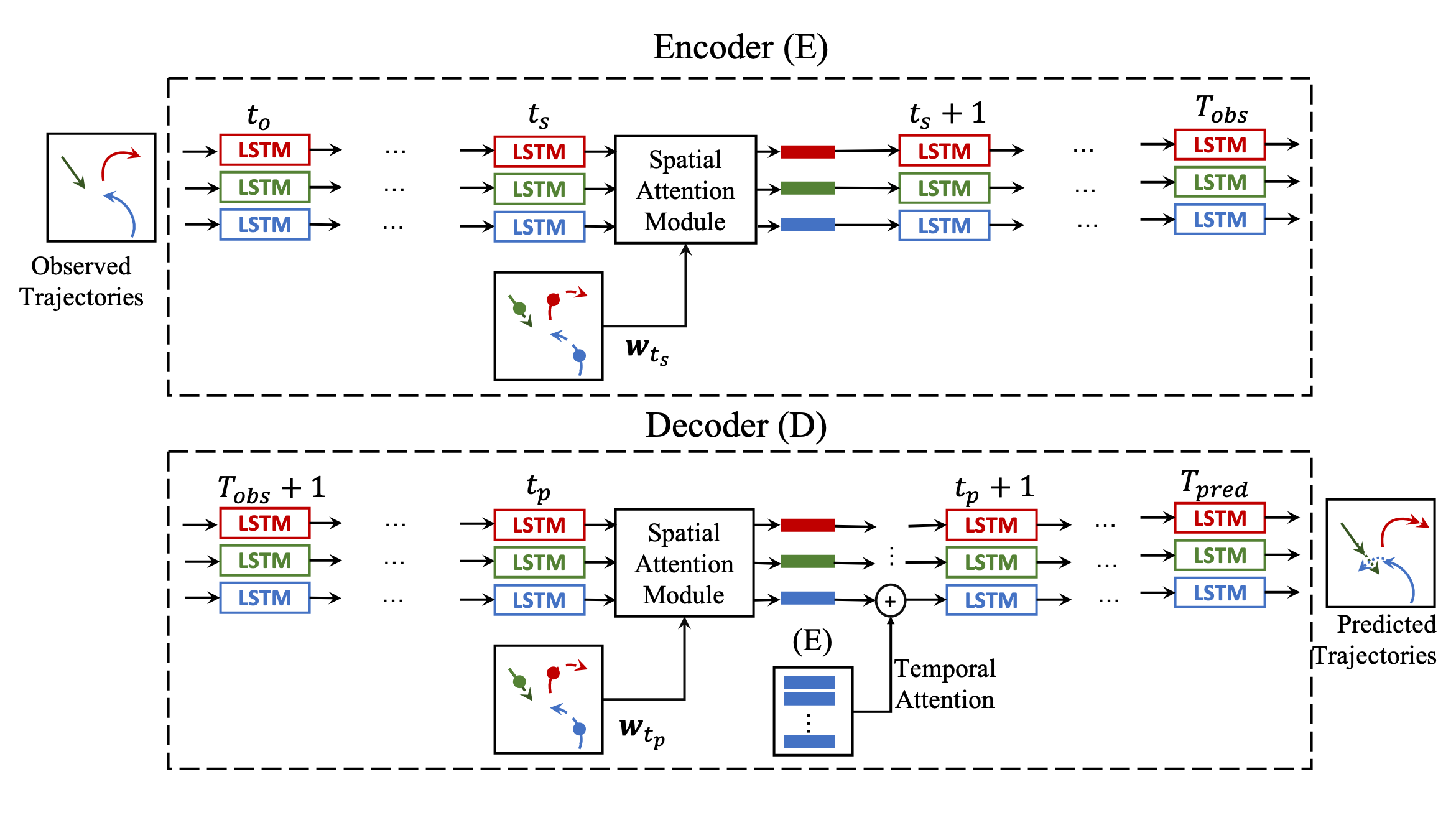}
\caption{SCAN Architecture. $\mathbf{w}_{t_s}$ represents the vector of spatial weights computed for each neighbor with respect to each pedestrian using Equation \ref{eq:hw_attn} for $t_s \in [t_0, T_{obs}]$, similarly $\mathbf{w}_{t_p}$ for $t_p \in [T_{obs}+1, T_{pred}]$ . In the decoding stage, temporal attention is interleaved with the spatial attention mechanism to enable the model to attend to observed spatial contexts.}
\label{fig:model}
\end{figure}
\textbf{Model Architecture.} At a high level, \textbf{SCAN} is an LSTM-based encoder-decoder framework. The encoder encodes each pedestrian’s observed trajectory into a fixed-length vector, and the decoder uses this fixed-length vector to predict each pedestrian’s predicted trajectory. 
Our proposed model architecture is shown in Figure \ref{fig:model}. We denote the number of pedestrians in the scene as $N$, observation time steps as $t_s \in [t_0, T_{obs}]$ and the prediction timesteps as $t_p \in [T_{obs}+1, T_{pred}]$. At a certain timestep $t$, we denote the trajectory of a pedestrian $p$, $p \in [1, N]$, by $\mathbf{x}^{p}_{t}$. Conventionally, the hidden state of an LSTM associated with modeling the trajectory of $p$ is updated using its hidden state at previous time step $t-1$, $h^{p}_{t-1}$ and $\mathbf{x}^{p}_{t}$. However, this update mechanism does not account for the spatial influences of other pedestrians on $p$'s trajectory.  

To take this spatial interaction into account, we incorporate a spatial attention mechanism, which will be explained in detail momentarily. 
Using this attention mechanism, the LSTM is able to incorporate spatial context experienced by $p$ by computing a spatially weighted hidden state, $\tilde{h}^{p}_{t_{s}-1}$. The LSTM then uses this spatially-weighted hidden state to compute the next hidden state for pedestrian $p$ using the conventional update mechanism:
\begin{equation}\label{eqn:lstm}
    h^{p}_{t} = \mathbf{LSTM}(\mathbf{x}^{p}_{t-1}, \tilde{h}^{p}_{t-1})
\end{equation}

This update mechanism is followed by both the LSTM encoder and LSTM decoder in our framework. By doing so, our framework is not only able to account for spatial influences that were experienced by $p$ in the observed trajectory, but also \emph{anticipate} the spatial influence of neighboring pedestrians on the trajectory of $p$ in the future. 
Using spatial attention in the prediction time window is similar to a pedestrian altering their path if they anticipate collision with another pedestrian at a future time step. 

While navigating through crowds, the spatial influence of neighbors causes pedestrians to temporarily digress from their intended trajectory to evade collision, respect personal space, etc. Therefore, while predicting intent for these pedestrians, some observed timesteps would be more reflective of their intent than others based on the spatial context associated with each observed timestep, $t_s$. In typical attention-based LSTM encoder-decoder frameworks, temporal attention is incorporated to enable the decoder to variably attend to the encoded hidden states. In our approach, we attempt to adopt temporal attention to enable our framework to attend to encoded \emph{spatial contexts}. 

At every $t_p \in [T_{obs+1}, T_{pred}]$, for a pedestrian $p$, the decoder attends to every spatially weighted hidden state, $\tilde{h}^{p}_{t_{s}}$, where $t_s \in [t_0, T_{obs}]$. To do so, the decoder compares the current spatially weighted hidden state for $p$, $\tilde{h}^{p}_{t_{p}}$ with all $\tilde{h}^{p}_{t_{s}}$, $t_s \in [t_0, T_{obs}]$ and assigns a score of similarity to each. The model then \emph{attends} more to the spatially weighted hidden states that have been assigned a higher score than others. This mechanism of attending variably to different time steps from the observation window is called temporal attention or soft attention~\cite{luong-etal-2015-effective}. In our model, we use the dot product as the scoring mechanism for temporal attention. Therefore, the score assigned to a $\tilde{h}^{p}_{t_s}$ would be maximum when $\tilde{h}^{p}_{t_s}$ = $\tilde{h}^{p}_{t_p}$, which would mean that the spatial context at $t_p$ is similar to an observed spatial context at $t_s$. 
Therefore, in our framework, \textbf{SCAN}, the decoder possesses a novel \emph{interleaved} spatially and temporally attentive architecture, that not only accounts for previous spatial interactions, but also accounts for the anticipated spatial interactions in the future, their influence on the pedestrian's intent thereof, and the variable influence of observed spatial contexts on the pedestrian's intent.  


%% file: spatial_attention_mechanism.tex
\textbf{Spatial Attention Mechanism.} As mentioned earlier, a pedestrian's intent is influenced by other pedestrians' trajectories and their expected intent. However, not all other pedestrians in a scene are of importance towards predicting the intent of a pedestrian. People navigating far off or towards different directions and not in line of sight of the pedestrian would have little to no effect on the pedestrian's intent. Therefore, to be able to understand and model spatial interactions experienced by a pedestrian, it is important to understand what the \emph{neighborhood} of the pedestrian is, i.e., the neighbors that have a spatial influence on the pedestrian. As discussed earlier, prior approaches have either made significant assumptions about this \emph{neighborhood}~\cite{Alahi_2016_CVPR}, assumed identical influence of all neighbors within this neighborhood irrespective of their orientations~\cite{Alahi_2016_CVPR, gupta2018social} or only used features such as distance from the pedestrian~\cite{Sadeghian_2019_CVPR}. Others, such as graph-based approaches~\cite{Mohamed_2020_CVPR, Kosaraju2019SocialBiGATMT, Vemula2018SocialAM} require learning a `weight' for all pairs of pedestrians in the scene. 

We introduce a concept called pedestrian domain, borrowed from an identical concept in ship navigation~\cite{pietrzykowski_uriasz_2009}. We define the domain of a pedestrian as the boundary of the area around a pedestrian, the intrusion of which by a neighbor causes the neighbor's trajectory to influence the intent of the pedestrian. Any other pedestrian that is beyond this boundary from the pedestrian of interest has no influence on the pedestrian's trajectory. Hereafter, we denote the domain by $\mathbf{S}$. The magnitude of influence of a neighbor, $p_2$, on that of a pedestrian of interest, $p_1$ at a certain instant $t$ is largely dependent on three factors: distance between the $p_1$ and $p_2$, $d^{21}_{t}$, relative bearing of $p_2$ from $p_1$ $\theta^{21}_{t}$, relative heading of $p_2$ to $p_1$, $\phi^{21}_{t}$. The influence of $p_2$ on the intent of $p_1$ at $t+1$ is then determined by computing its spatial weight or \textit{score} at $t$:
\begin{equation}\label{eq:hw_attn}
       \mathbf{score}(p_1, p_2)_{t} =  w^{21}_t = \mathbf{ReLU}(\mathbf{S}^{\theta^{21}_t,\phi^{21}_t} - d^{21}_t)
\end{equation}
where $\mathbf{S}^{\theta^{21}_t,\phi^{21}_t}$ denotes the value of the pedestrian domain $\mathbf{S}$ for $\theta^{21}_t$,$\phi^{21}_t$. 
Imagine if we discretized bearing and heading such that any encounter between agents can be put in a ``bin''. Let $\mathbf{S}\in {R}^{m,n}$, where the set $i\in\left\{1,\dots,m \right\}$ (or $j\in\left\{1,\dots,n \right\}$) maps to an interval in the relative bearing $[(i-1)\cdot\alpha,i\cdot\alpha)$ where $\alpha = \frac{360^\circ}{m}$ (similar reasoning for heading). At the risk of overloading notation, we define $\mathbf{S}^{\theta^{21}_t,\phi^{21}_t}$ to be the element $s_{i,j}$ of $\mathbf{S}$ such that the encounter geometry is a kind of indicator function for the appropriate index on $i,j$.
For example, if both bearing and heading are discretized at $30^\circ$ ($m=n=12$) increments and an encounter occurs at time $t=0$ of $\theta^{21}_0=5^\circ$ and $\phi^{21}_0=185^\circ$ (potentially a collision course, by the way) it will lead to learning of the domain $\mathbf{S}$ in the increment of $\theta^{21}_0\in[0,30)\wedge\phi^{21}_0\in[180,210)$, or in this case $\mathbf{S}^{{\theta^{21}_0,\phi^{21}_0}}$ maps to the element $s_{1,7}$ of $\mathbf{S}$.
\begin{figure}
    \centering
    \includegraphics[width=0.9\linewidth]{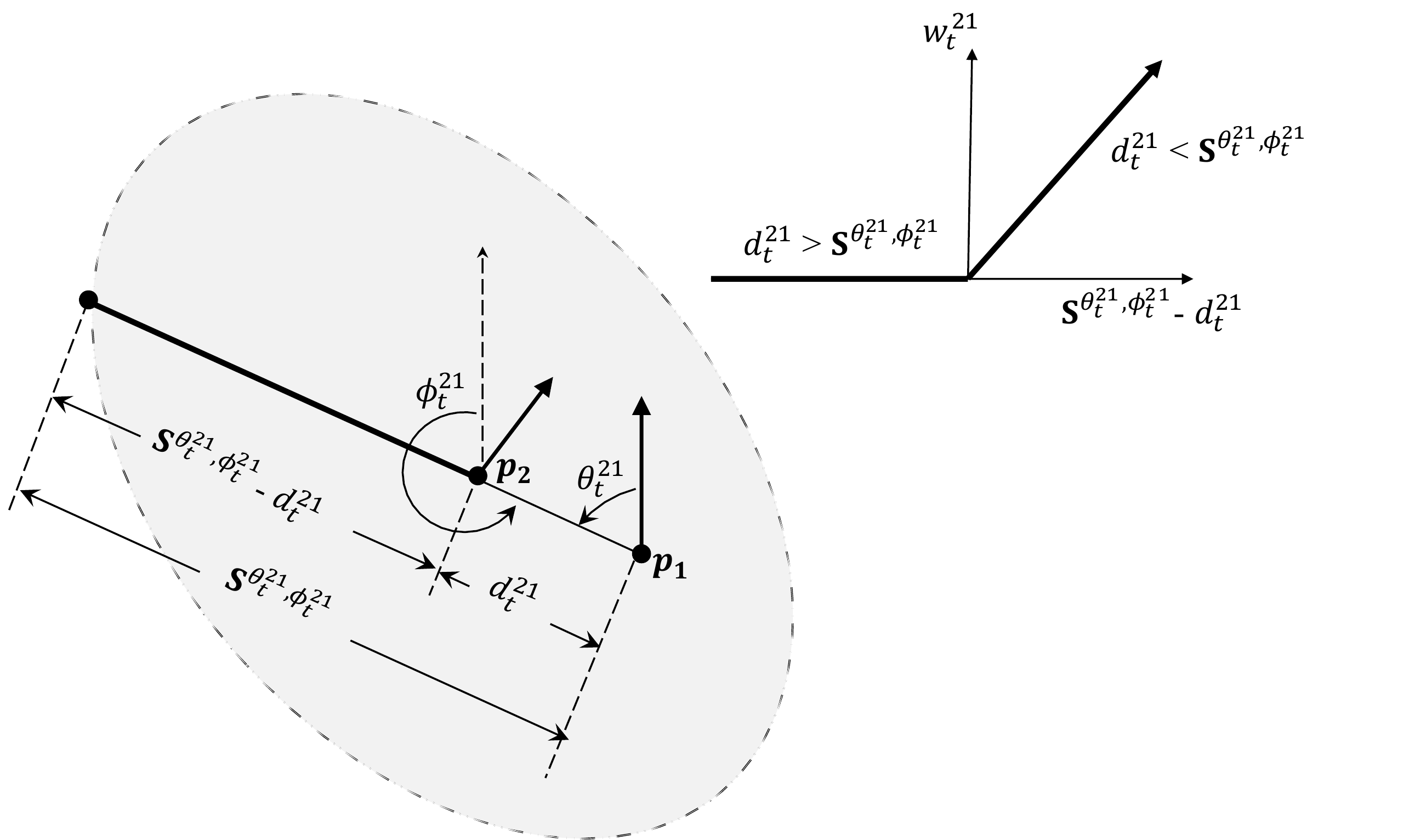}
    \caption{For $p_2$ at distance $d^{21}_{t}$, relative bearing $\theta^{21}_{t}$, and relative heading $\phi^{21}_{t}$ from $p_1$, the spatial weight or score of $p_2$ at $t$ increases with increase in distance from $\mathbf{S}^{\theta^{21}_{t},\phi^{21}_t}$.}
    \label{fig:my_label}
\end{figure}

This weighting mechanism directly translates into a pedestrian closer to the self, and hence farther from $\mathbf{S}$ having a larger weight, and hence a larger influence on the self. Similarly, a pedestrian closer to the boundary, $\mathbf{S}$, and hence farther from the self would have a smaller influence on the self. The activation function $\mathbf{ReLU}$ ensures that if a pedestrian $p_2$ is at a distance $d^{21}_{t} \geq \mathbf{S}$ from $p_1$ at $t$, its influence on the intent of $p_1$ at $t+1$ is 0. This allows the model to determine the domain as an area beyond which another pedestrian in the scene would not affect the self and vice-versa. 

However, using this spatial scoring mechanism, a neighbor at a certain distance and orientation with respect to the pedestrian of interest would always have the same spatial influence on the pedestrian's trajectory, irrespective of crowd densities. However, a certain neighbor $p_2$ at a (large) distance from $p_1$, with a small positive value for $w^{21}_t$ might not affect $p_1$ much in a densely crowded setting but might influence $p_1$ more in a sparsely crowded environment. Simply put, while navigating in environments that are not too crowded humans often tend to change their trajectories as a response to someone that is relatively far away; however, in crowded settings, the same neighbor at the same orientation and distance does not pose an immediate risk of collision and hence does not influence the pedestrian's trajectory as much. To account for this varying spatial influence as a result of varying crowd densities, we normalize the scores for all neighbors for each pedestrian in the frame, 
\begin{equation}\label{eqn:spatial_alignment}
   \mathbf{score}(p_1, p_2)_{t} = \frac{\mathbf{exp}(\mathbf{score}(p_1, p_2)_{t})}{\sum_{n\in N'}\mathbf{exp}(\mathbf{score}(p_1, p_n)_{t}))} 
\end{equation}
where $n\in{N'}$ are all pedestrians in the frame apart from the pedestrian of interest, $p_1$. Once the spatial scores have been computed for every pair of pedestrians, 
we compute a \emph{spatial context vector} for each pedestrian that represents the \emph{spatial context} experienced from the pedestrian's perspective at $t$. For example, the spatial context vector for $p_1$ at $t$ is computed as, 
\begin{equation}
   \tilde{\mathbf{C}}^{p_1}_{t} = \sum_{n\in N'} \mathbf{score}(p_1, p_n)_th^{p_n}_{t} 
\end{equation}
This spatial context vector contains meaningful information about the spatial orientation of other pedestrians in the frame from $p_1$'s perspective at $t$ and hence the amount of knowledge (hidden states) shared with $p_1$ about its neighbors depending on their orientations.  This spatial context is then concatenated with the hidden state of the pedestrian at $t$ before it is fed to the LSTM. For $p_1$,
\begin{equation}
    \tilde{h}^{p_1}_{t} = \mathbf{concat}(h^{p_1}_t, \tilde{C}^{p_1}_t)
\end{equation}
This gives the model relevant information of both the pedestrian's own hidden state as well as spatial context from the pedestrian's perspective. Every pedestrian in the frame has a unique spatial context, which is the spatial orientation and influence of neighbors experienced by the pedestrian at $t$ from its own perspective instead of a global perspective. 


%% file: multimodalbehavior.tex
\textbf{Multiple Socially Plausible Paths.} Given an observed trajectory, there can be more than one \emph{socially plausible} trajectory that a pedestrian can take in the future. A socially plausible trajectory would account for spatial influence of neighboring pedestrians' trajectories and respect social norms. For safe navigation, it is imperative to be able to account for the fuzzy nature of human motion and be able to generate multiple socially plausible future trajectories instead of narrowing down on one average expected behavior. To do so, we leverage the generative modeling abilities of GANs (Generative Adversarial Networks)~\cite{10.5555/2969033.2969125}. Briefly, the training process of GANs is formulated as a two player min-max game between a generator and a discriminator. The generator generates candidate predictions and the discriminator evaluates them and scores them as real/fake. In our case, the goal of the generator is to be able to generate predictions that are consistent with the observed trajectory and are also consistent with the observed and intended spatial contexts, hence \emph{socially plausible}. The discriminator must be able to discern which trajectories are real, and which are generated. GANs have also been previously adopted for pedestrian intent prediction~\cite{gupta2018social, Sadeghian_2019_CVPR, Kosaraju2019SocialBiGATMT, Vemula2018SocialAM}. 

\textbf{\textit{Generator.}} The generator of our model is basically the encoder-decoder framework that we described above. The goal of generator is to learn how to generate realistic trajectories that are consistent with the observed trajectories and the observed spatial contexts that are incorporated in the encoded representation of each pedestrian by virtue of the interleaved spatial attention mechanism. We achieve this by initializing the hidden state of the decoder for a pedestrian, $p$, as
\begin{equation}
    h^{p}_{T_{obs+1}} = [h^{p}_{T_{obs}}, z]
\end{equation}
where $z$ is a noise vector, sampled from $\mathcal{N}(0,1)$ and $h^{p}_{T_{obs}}$ is the encoded representation for pedestrian, $p$, or the final hidden state of the LSTM encoder pertaining to $p$. A difference of our approach in comparison to prior multimodal intent forecasting approaches is that in addition to the pedestrian's encoding, they also condition the generation of output trajectories on social context vectors~\cite{gupta2018social} that summarise the spatial context of the pedestrian, $p$. In our framework, our interleaved spatial attention mechanism already accounts for spatial context in the encoded representation.

\textbf{\textit{Discriminator.}} The discriminator contains a separate encoder. This encoder takes as input the $N$ `ground truth' trajectories over $[t_0, T_{obs}]$ and the $N$ generated trajectories over $[t_0, T_{obs}]$ and classifies them as `real' or `fake'. The encoder in the discriminator also uses the spatial
attention mechanism at each time step, therefore ideally the goal of the discriminator is to classify the trajectories as real/fake while taking into account social interaction rules. This would imply that trajectories that do not seem to comply with social navigation norms and hence are not socially plausible would be classified as fake.

%% file: experiments.tex



 \textbf{Datasets.}
 We evaluate \textbf{SCAN} on two publicly available pedestrian-trajectory datasets: ETH\cite{5459260} and UCY\cite{Lerner2007CrowdsBE}. The datasets contain birds eye-view frames sampled at 2.5 fps and 2D locations of pedestrians walking in crowded scenes. The ETH dataset contains two sub-datasets (annotated ETH and HOTEL) from two scenes, each with 750 pedestrians. The UCY dataset contains two scenes with 786 pedestrians, split into three sub-datasets (ZARA1, ZARA2, UNIV). These datasets contain annotated trajectories of pedestrians interacting in several social situations and include challenging behavior such as collision avoidance, movement in groups, yielding right of way, couples walking together, groups crossing groups, etc.\cite{5459260}.

\textbf{Baselines.} 
We compare our model against several baselines:
\begin{enumerate*}[label=(\alph*)]
    \item \textit{Linear:} A linear regressor with parameters estimated by minimizing least square error; \item \textit{LSTM:} An LSTM that models only individual pedestrian trajectory without accounting for any spatial interactions;
    \item \textit{Social LSTM}~\cite{Alahi_2016_CVPR}: Uses a pooling mechanism to model spatial influence of neighbors within an assumed spatial grid and models each pedestrian's trajectory using an LSTM;
    \item \textit{S-GAN}~\cite{gupta2018social}: Models spatial interactions using a grid-based pooling mechanism, and models each pedestrian's trajectory using a GAN-based framework similar to ours;
    \item \textit{S-GAN-P}~\cite{gupta2018social}: Similar framework to \textit{S-GAN}, but incorporates their proposed pooling mechanism to model spatial interactions;
    \item \textit{SoPhie GAN}~\cite{Sadeghian_2019_CVPR}: Models agent trajectories using a LSTM-GAN framework with additional modules to incorporate social attention and physical scene context;
    \item \textit{Social Attention}~\cite{Vemula2018SocialAM}: Models pedestrian trajectory prediction as a spatio-temporal graph, also incorporates features like relative orientation and distances in the spatial edges of the graph;
    \item \textit{Social Ways}~\cite{amirian2019social}: GAN-based framework that also incorporates relative orientation features as a prior over the attention pooling mechanism;
    \item \textit{Social-BiGAT}~\cite{Kosaraju2019SocialBiGATMT}: Graph-based GAN that uses a graph attention network (GAT) to model spatial interactions and an adversarially trained recurrent encoder-decoder architecture to model trajectories;
    \item \textit{Trajectron}~\cite{ivanovic2018trajectron}:  An LSTM-CVAE encoder-decoder which is explicitly constructed to match the spatio-temporal structure of the scene; and 
    \item \textit{Trajectron++}~\cite{SalzmannIvanovicEtAl2020}: Similar to ~\cite{ivanovic2018trajectron}, but uses directed edges in the spatio-temporal graph modeling the scene.
\end{enumerate*}

\textbf{Implementation.}
We follow a leave-one-out evaluation methodology to train and test \textbf{SCAN} on each of the five datasets, training on four datasets and testing on the fifth. As with all prior approaches, we observe the trajectory for 8 time steps (2.8 seconds) and predict intent over future 12 time steps (3.2 seconds). Model parameters are iteratively trained using Adam optimizer with a batch size of 32 and an initial learning rate of 0.001. The model is implemented in PyTorch and trained using a single GPU. In both the encoder and the decoder, the positional information pertaining to each pedestrian in the frame is first embedded into 16 dimensional vectors using a linear layer. The hidden states for both the encoder and the decoder LSTMs are 32 dimensional vectors. In the decoder, a linear layer is used to convert the LSTM output to the  ($x$,$y$) coordinates predicted for the pedestrians. Relative bearing and relative heading are discretized at $30^{o}$.

\input{results}

%% file: results.tex
\textbf{Quantitative Comparison.}
We compare two versions of our model - \textbf{SCAN}, the proposed encoder-decoder framework with interleaved spatial and temporal attention, and \textbf{vanillaSCAN}, the proposed encoder-decoder architecture sans the temporal attention in the decoder - with the deterministic baselines (\textit{Linear}, \textit{Social LSTM}~\cite{Alahi_2016_CVPR}, \textit{Social Attention}~\cite{Vemula2018SocialAM}, deterministic \textit{Trajectron++}~\cite{SalzmannIvanovicEtAl2020}) in Table ~\ref{tab:table1}. We also compare GAN-based generative framework, \textbf{generativeSCAN} with the generative baselines (\textit{S-GAN}~\cite{gupta2018social}, \textit{S-GAN-P}~\cite{gupta2018social}, \textit{SoPhie GAN}~\cite{Sadeghian_2019_CVPR}, \textit{Social Ways}~\cite{amirian2019social}, \textit{Trajectron}~\cite{ivanovic2018trajectron}, generative \textit{Trajectron++}~\cite{SalzmannIvanovicEtAl2020}) in Table \ref{tab:table2}. 
We report our results using two metrics: \textit{Average Displacement Error (ADE)}, which is the average L2 distance between ground truth trajectories and predicted trajectories over all predicted time steps, and \textit{Final Displacement Error (FDE)}, which is the average displacement error between final predicted destination of all pedestrians at the end of the time window and the true final destination at $T_{pred}$. 
\input{table1}
\input{table2}
In Table~\ref{tab:table1}, while we mention results for \emph{Social Attention}~\cite{Vemula2018SocialAM}, as are reported in their paper, it is not directly comparable to our method because, as mentioned in their paper, they process their dataset differently in comparison to the other baselines (and our method).  While \emph{Trajectron++}~\cite{SalzmannIvanovicEtAl2020} has an average lower ADE, \textbf{SCAN} has a lower final displacement error, implying that its ability to anticipate spatial interactions into the future enable it to predict a more accurate final destination. Both \textbf{vanillaSCAN} and \textbf{SCAN} are largely able to outperform the other deterministic baselines on the five datasets. Interleaving temporal attention with spatial attention in \textbf{SCAN} also enables the model to capture long-term or high-level intent more accurately, which is reflected in its lower FDE values compared to \textbf{vanillaSCAN}. In Table~\ref{tab:table2}, we compare \textbf{generativeSCAN} with other baselines that account for multimodal pedestrian behavior. \emph{Sophie GAN}~\cite{Sadeghian_2019_CVPR} takes into account physical scene information while making trajectory predictions. Despite our model being agnostic to such information, it is able to achieve lower ADE and FDE than both \emph{Sophie GAN} and \emph{S-GAN}~\cite{gupta2018social}. Our model is also able to outperform \emph{Social-Ways} on both the Zara datasets. \emph{Social-BiGAT}~\cite{Kosaraju2019SocialBiGATMT}, which uses a graph attention network~\cite{Velickovic2018GraphAN} to model spatial influences, is able to slightly outperform our model on an average. As we explain later, our spatial attention mechanism in fact outperforms a graph-based attention mechanism for modeling spatial influences, hence \emph{Social-BiGAT}'s performance may be attributed to its ability to also include scene information while making its predictions. \emph{Trajectron++} is largely able to outperform \textbf{generativeSCAN} across all five datasets. While it simply uses a directed spatiotemporal graph to model agent interactions, \emph{Trajectron++}~\cite{ivanovic2018trajectron} incorporates a conditional variational autoencoder (CVAE)~\cite{NIPS2015_5775} to sample multimodal trajectories conditioned on future behavior, as opposed to \textbf{generativeSCAN} and other baselines that are GAN-based. \\
\textbf{Variety loss and diversity loss.}
While accounting for multimodal pedestrian behavior, it is important to ensure that the generated predictions are diverse and not simply multiple `close to average' predictions. We train \textbf{generativeSCAN} using adversarial loss and L2 loss. However, while the trained model is able to generate multiple socially plausible trajectories, these are largely very similar predictions. To encourage diversity in generated trajectories, we adopt \emph{variety loss}, as proposed in~\cite{gupta2018social}. For each scene, the generator generates $k$ possible output predictions by randomly sampling $z$ from $\mathcal{N}(0,1)$ and penalizing the `best prediction', i.e., the one with the least ADE. However, training the model with a large $k$ value is computationally expensive because it involves $k$ forward passes per batch in the training dataset. Further, we observed that increasing $k$ does not improve the diversity of the generated trajectories substantially. Therefore, we incorporate another loss function, \emph{diversity loss}, which essentially penalizes the generator for generating similar trajectories. For $N$ pedestrians in the frame, 
\begin{equation}
    \mathcal{L}_{diversity} = \frac{1}{N}\sum_{i,j \in k} \mathrm{exp}(-d_{ij})
\end{equation}
where $d_{ij}$ is the average euclidean distance between trajectories $i$ and $j$. 
The generator is then trained using the sum of adversarial loss, variety loss and the diversity loss weighted by parameter $\lambda$. 
In Figure~\ref{fig:diversity}, we analyze the effect of increasing $k$ and increasing $\lambda$ on the diversity in generated trajectories in a crossing scenario. More diverse trajectories can be generated by increasing $\lambda$ value for a smaller $k$ value. \\
\input{table3}
\textbf{Modeling Spatial Interactions as a Graph.} Our spatial attention mechanism has certain similarities to graph attention networks~\cite{Velickovic2018GraphAN}, since we initially consider all nodes (pedestrians) to be connected, or influence each other, and then proceed to learn the `domain' which enables us to learn these influences or edges during training. While we compare our model's performance against a number of graph attention based trajectory forecasting methods in Table \ref{tab:table2}, to validate the performance benefits of our proposed spatial attention mechanism, we also evaluate an ablation that uses a graph attention network (GAT) in place of our spatial attention mechanism in \textbf{SCAN} with the rest of the framework being the same. The results are reported in Table \ref{tab:table3}. Computationally, both mechanisms are nearly the same. The slight overhead for our method comes from having to compute distance, bearing, heading for each prediction time step in order to compute spatial attention weights. Our spatial attention mechanism is therefore capable of achieving lower error in comparison to a graph attention network, while being comparably fast at inference time. Further, the learned domain parameter informs interpretability of the model's predictions since it provides information about the neighbors that influences the pedestrian and its intent. 
\input{qualitative_analysis}

%% file: table1.tex
\begin{table*}
{\renewcommand{\arraystretch}{1}
    \centering
    \small 
    \begin{tabular}{|c|c|c|c|c|c||c|c|} \hline 
         \multirow{2}{*}{\textbf{Dataset}} & \multicolumn{7}{c|}{ \textbf{ADE / FDE (m)}} \\ \cline{2-8}
         & \textit{Linear} & \textit{LSTM} & \textit{Social LSTM} & \textit{Social Attention}  & \textit{Trajectron++} & \textbf{vanillaSCAN} & \textbf{SCAN} \\ \hline \hline 
         ETH & 1.33 / 2.94 & 1.09 / 2.41 & 1.09 / 2.35 & 0.39 / 3.74 & 0.71 / 1.66 & 0.79 / 1.36 & 0.78 / 1.29 \\ 
         Hotel & 0.39 / 0.72 & 0.86 / 1.91 & 0.79 / 1.76 & 0.29 / 2.64 & 0.22 / 0.46 & 0.46 / 0.95 & 0.40 / 0.76  \\ 
         Univ & 0.82 / 1.59 & 0.61 / 1.31 & 0.67 / 1.40 & 0.33 / 3.92 & 0.41 / 1.07 & 0.64 / 1.34 & 0.62 / 1.28 \\ 
         Zara1 & 0.62 / 1.21 & 0.41 / 0.88 & 0.47 / 1.00 & 0.20 / 0.52 & 0.30 / 0.77 & 0.39 / 0.86 & 0.38 / 0.80 \\ 
         Zara2 & 0.77 / 1.48 & 0.52 / 1.11 & 0.56 / 1.17 & 0.30 / 2.13 & 0.23 / 0.59 & 0.33 / 0.71 & 0.33 / 0.72 \\ \hline \hline 
         Average & 0.79 / 1.59 & 0.70 / 1.52 & 0.72 / 1.54 & 0.30 / 2.59 & 0.37 / 0.91 & 0.52 / 1.04 & 0.50 / 0.97 \\ \hline 
    \end{tabular}
    \caption{Comparison of our models, vanillaSCAN and SCAN against other deterministic baselines.}
    \label{tab:table1}}
\end{table*}

%% file: table2.tex
\begin{table*}
{\renewcommand{\arraystretch}{1}
    \centering
    \small
    \begin{tabular}{|p{0.05\linewidth}|c|c|c|p{0.1\linewidth}|c|p{0.1\linewidth}||c|} \hline 
         \multirow{2}{*}{\textbf{Dataset}} & \multicolumn{7}{c|}{ \textbf{ADE / FDE (m)}, Best of 20 } \\ \cline{2-8}
         & \textit{S-GAN} & \textit{Sophie GAN} & \textit{Social Ways} & \textit{Social BiGAT}  & \textit{Trajectron} & \textit{Trajectron++} & \textbf{generativeSCAN} \\ \hline \hline 
         ETH & 0.81 / 1.52 & 0.70 / 1.43 & 0.39 / 0.64 & 0.69 / 1.29  & 0.59 / 1.14 & 0.39 / 0.83 &  0.79 / 1.49 \\ 
         Hotel & 0.72 / 1.61 & 0.76 / 1.67 & 0.39 / 0.66 & 0.49 / 1.01 &  0.35 / 0.66 & 0.12 / 0.19 & 0.37 / 0.74 \\ 
         Univ & 0.60 / 1.26 & 0.54 / 1.24 & 0.55 / 1.31 & 0.55 / 1.32 &  0.54 / 1.13 &  0.20 / 0.44 & 0.58 / 1.23 \\ 
         Zara1 & 0.34 / 0.69 & 0.30 / 0.63 & 0.44 / 0.64 & 0.30 / 0.62&  0.43 / 0.83& 0.15 / 0.32 & 0.37 / 0.78 \\ 
         Zara2 & 0.42 / 0.84 & 0.38 / 0.78 & 0.51 / 0.92 & 0.36 / 0.75& 0.43 / 0.85  & 0.11 / 0.25 & 0.31 / 0.66 \\ \hline \hline 
         Average & 0.58 / 1.18 & 0.54 / 1.15 & 0.46 / 0.83 &  0.48 / 1.00 & 0.56 / 1.14& 0.19 / 0.41 & 0.48 / 0.98 \\ \hline 
    \end{tabular}
    \caption{Comparison of our generative model, generativeSCAN with other generative baselines. The results reported for all generative models are `best of 20', which means the ADE for the trajectory with least ADE out of 20 generated trajectories per sample is reported. The FDE value is reported for the trajectory with the best ADE.}
    \label{tab:table2}}
\end{table*}

%% file: table3.tex
\begin{table}
    \centering
    \small
    {\renewcommand{\arraystretch}{1}
    \begin{tabular}{|c|p{0.2\linewidth}|c||p{0.1\linewidth}|p{0.1\linewidth}|} \hline 
        \multirow{2}{*}{\textbf{Dataset}} & \multicolumn{2}{c|}{\textbf{ADE / FDE (m)}} & \multicolumn{2}{c|}{\textbf{Time (s)}} \\ \cline{2-5}
         & \textbf{Graph based SCAN} & \textbf{SCAN} & \textbf{Graph based SCAN} & \textbf{SCAN} \\ \hline \hline 
         ETH & 0.99 / 1.90 & 0.78 / 1.29 & 0.068 & 0.076 \\ 
         Hotel & 0.53 / 1.19 & 0.40 / 0.76 & 0.071 & 0.074\\ 
         Univ & 0.79 / 1.46 & 0.62 / 1.28 & 0.074 & 0.077 \\ 
         Zara1 & 0.43 / 0.92 & 0.38 / 0.80 & 0.074 & 0.079\\ 
         Zara2 & 0.41 / 0.88 & 0.33 / 0.72 & 0.071 & 0.078 \\ \hline \hline 
         Average &  0.63 / 1.27 &  0.50 / 0.97 & 0.072 & 0.077 \\ \hline 
    \end{tabular}
    \caption{Quantitative comparison of SCAN with Graph based SCAN, an ablation that models spatial influence using graph attention networks (GATs). The inference time reported is averaged across ten evaluation runs.}
    \label{tab:table3}}
\end{table}

%% file: qualitative_analysis.tex
\input{fig_diversity}

\input{table4}
\input{table5}
\input{fig_discretization}
{\textbf{Collision Analysis.}} 
To demonstrate the capability of our spatial attention mechanism to predict safe, \emph{socially acceptable} trajectories, we evaluate the ability of trajectories predicted by our model to avoid ``collisions". To do so, we calculate the average percentage of pedestrians \emph{near-collisions} across the five evaluation datasets. As in \cite{Sadeghian_2019_CVPR}, for a given scene, if the euclidean distance between any two pedestrians drops below 0.10 m, we say that a \textit{near-collision} has occurred. In Table ~\ref{tab:table4}, we compare the average percentage of colliding pedestrians for predictions generated by \textbf{SCAN} against several other baselines. Our model is able to predict much more socially acceptable trajectories in comparison to other baselines. Further, the average percentage of colliding pedestrians per frame for each dataset as obtained by our model's predictions is much closer to the ground truth as compared to the other baselines. \textit{Social-GAN}~\cite{gupta2018social} uses a pooling mechanism to incorporate spatial influences of neighboring pedestrians, and \textit{Sophie-GAN} uses a sorting mechanism to incorporate distances while taking spatial influences into account. Further, \textit{Sophie-GAN}~\cite{Sadeghian_2019_CVPR} also incorporates scene context towards making more informed predictions. From Table \ref{tab:table4}, we can conclude that our proposed spatial attention mechanism is not only able to generate more socially acceptable trajectories, but is also able to capture the social behavior in the ground truth trajectories. \\
{\textbf{Effect of Different Bearing, Heading. Discretizations.}} 
In order to learn the pedestrian domain $\mathbf{S}$, we discretize the space of relative bearing and relative heading values such that any encounter between agents can be put in a ``bin''. In our evaluation, we choose to discretize relative bearing, $\theta$ and relative heading, $\phi$ values into bins of $\Delta\theta= \Delta\phi=30^{o}$. Figure \ref{fig:discretization}a. plots the variation in test ADE on ZARA1 dataset with increasing $\Delta\theta=\Delta\phi$. A more fine-grained discretization than $30^{o}$ has a higher test ADE. Similarly, more coarse-grained discretizations lead to higher test ADE values. A discretization of $360^{o}$ would correspond to a uniform value of $\mathbf{S}$ irrespective of relative bearing and relative heading values of a neighbor. Figure \ref{fig:discretization}b. also plots the number of learnable parameters in $\mathbf{S}$ as a function of discretization values. As is true of deep learning based architectures in general, a highly parameterized domain and lower parameterized $\mathbf{S}$ domains do not generalize well to the test dataset. \\
{\textbf{Effect of Varying Prediction Horizon Lengths.}} 
In Table \ref{tab:table5} we observe the average displacement error (ADE) for \textbf{SCAN} across all five datasets against various prediction horizon lengths for the same observed time window. For ZARA1, ZARA2 and HOTEL, the increase in ADE as the prediction time window is increased from 12 to 20 timesteps is $\approx$ 0.2 m. Therefore, using the same observed time window of 8 timesteps, \textbf{SCAN} is able to predict longer trajectories fairly accurately.

%% file: fig_diversity.tex
\begin{figure}
    \centering
    \includegraphics[width=0.4\textwidth]{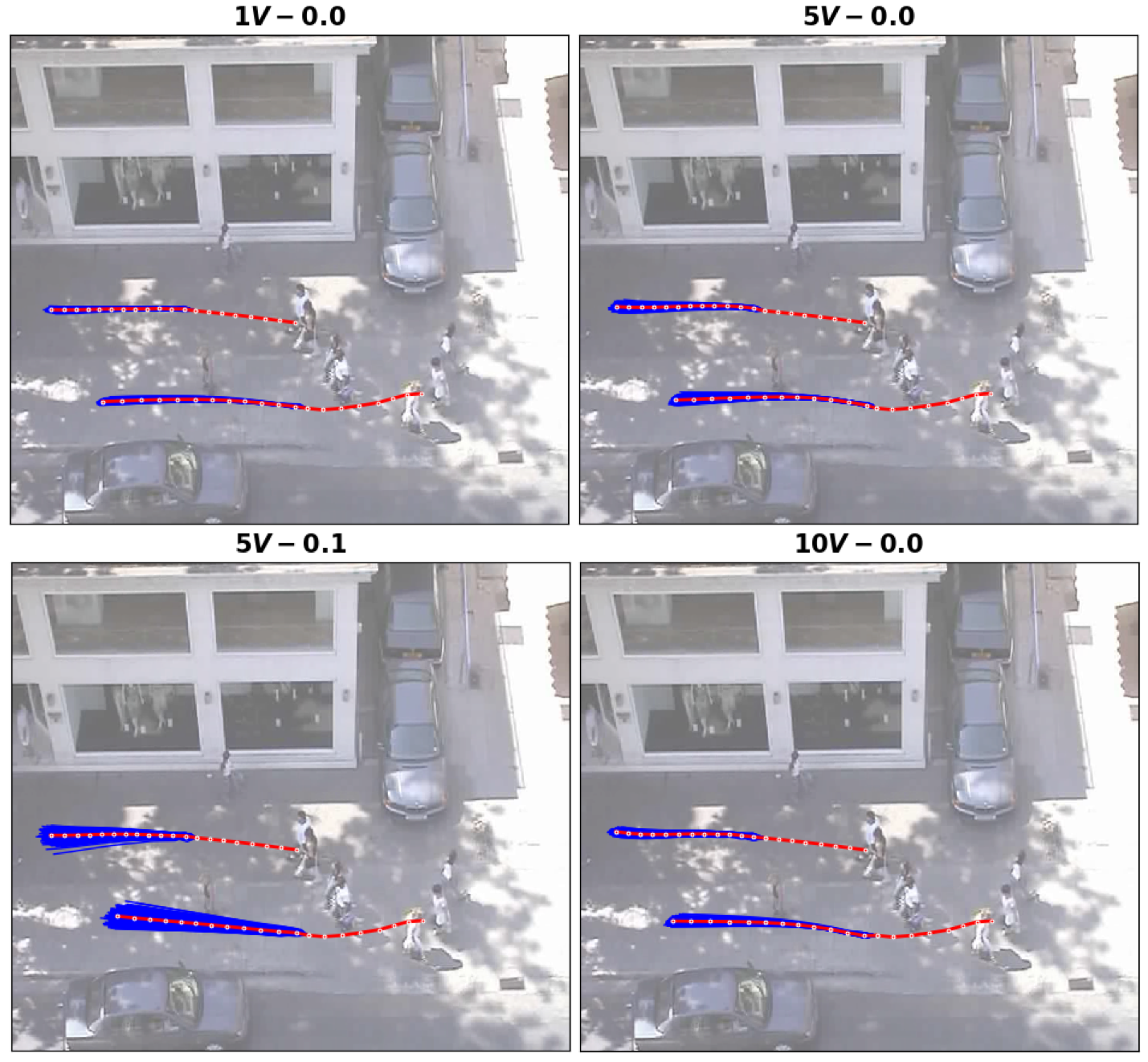}
    \caption{Effect of increasing $k$ and $\lambda$ on the diversity of generated trajectories on a scenario from Zara1 test dataset. We visualize 300 generated trajectories (in blue), and their mean (in red) for some pedestrians in the frame. Each plot is titled $k$V-$\lambda$.}
    \label{fig:diversity}
\end{figure}

%% file: table4.tex
\begin{table}
    \centering
    \small
    {\renewcommand{\arraystretch}{1}
    \begin{tabular}{|c||p{0.1\linewidth}|p{0.1\linewidth}|p{0.1\linewidth}|p{0.1\linewidth}|c|} \hline 
         Dataset & \textit{Ground Truth} & \textit{Linear} & \textit{Social-GAN} & \textit{SoPhie GAN} & \textbf{SCAN}  \\ \hline \hline 
         ETH & 0.000 & 3.137 & 2.509 & 1.757 & 0.793 \\
         Hotel & 0.092 & 1.568 & 1.752 & 1.936 & 1.126 \\
         Univ & 0.124 & 1.242 & 0.559 & 0.621 & 0.481 \\
         Zara1 & 0.000 & 3.776 & 1.749 & 1.027 & 0.852 \\ 
         Zara2 & 0.732 & 3.631 & 2.020 & 1.464 & 3.109 \\ \hline\hline  
         Average & 0.189 & 2.670 & 1.717 & 1.361 & 1.272 \\ \hline 
    \end{tabular}
    \caption{Average \% of collisions per frame for each of the five evaluation datasets.}
    \label{tab:table4}}
\end{table}

%% file: table5.tex
\begin{table}[t]
    \centering
    \small 
    {\renewcommand{\arraystretch}{1}
    \begin{tabular}{|c|c|c|c|} \hline 
       Dataset  & \textbf{8} & \textbf{12} & \textbf{20} \\ \hline \hline 
        ETH & 0.69 / 1.28 & 0.78 / 1.29 & 1.40 / 2.87 \\ 
        Hotel & 0.34 / 0.64 & 0.40 / 0.76 & 0.57 / 1.20 \\
        Univ & 0.36 / 0.74 & 0.62 / 1.28 & 1.57 / 3.11 \\
        Zara1 & 0.24 / 0.49 & 0.38 / 0.80 & 0.78 / 1.70 \\
        Zara2 & 0.20 / 0.43 & 0.33 / 0.72 & 0.58 / 1.32 \\ \hline \hline 
        Average & 0.37 / 0.72 & 0.50 / 0.97 & 0.98 / 2.04 \\ \hline 
    \end{tabular}
    \caption{ADE / FDE for SCAN on prediction lengths 8, 12 and 20.}
    \label{tab:table5}}
\end{table}

%% file: fig_discretization.tex
\begin{figure}[t]
    \centering
    \includegraphics[width=0.45\textwidth]{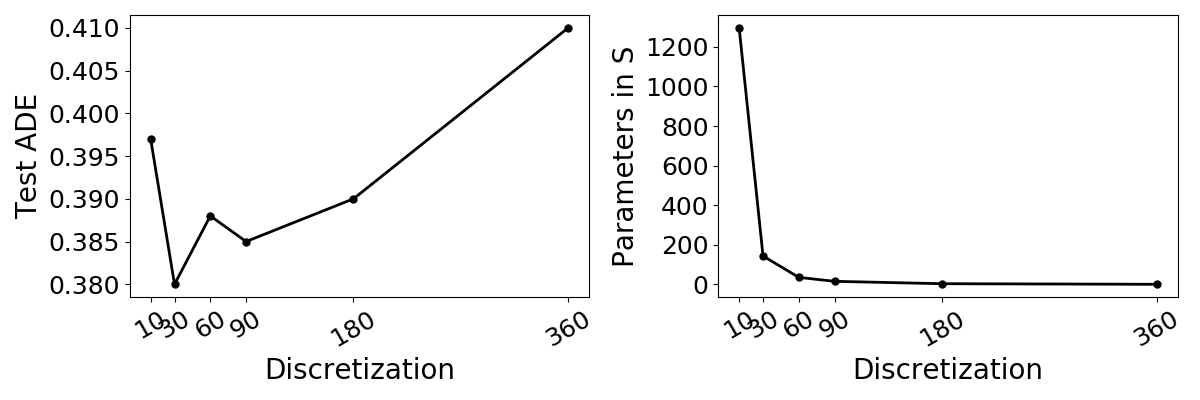}
    \caption{(a) Effect of increasing relative bearing, relative heading discretization on Test ADE, (b) Effect of increasing relative bearing, heading discretization on learnable parameters in the pedestrian domain, $\mathbf{S}$. }
    \label{fig:discretization}
\end{figure}

%% file: conclusion.tex
In this work, we propose \textbf{SCAN}, a novel trajectory prediction framework for predicting pedestrian intent. A key contribution of this work is the novel spatial attention mechanism, that is able to model spatial influence of neighboring pedestrians in a manner that is parameter efficient, relies on less assumptions and results in more accurate predictions. We also propose \textbf{generativeSCAN} that accounts for the multimodal nature of human motion and is able to predict multiple socially plausible trajectories per pedestrian in the scene. Despite being agnostic to scene context and relevant physical scene information, our model is able to match or even outperform existing baselines that use such information. This work can also be extended to predicting trajectories for heterogeneous agents with different trajectory dynamics. The spatial attention mechanism introduced in this work can be used to infer more domain-specific knowledge, such as the influence of different kinds of agents on each other (for example, the effect of a skateboarder on a cyclist's trajectory) and use these to either explain or inform model predictions.At a more fundamental level, \textbf{SCAN} is a general framework that can be applied to any sequence-to-sequence modeling application where cross-LSTM knowledge can help improve performance. This can include human action recognition~\cite{8428616,song2016end}, modeling human-object interactions~\cite{7780942, Qi2018LearningHI}, video classification~\cite{Wu2015ModelingSC}. An important advantage of \textbf{SCAN} is its ability to infer domain knowledge from the observation dataset and hence yield improved predictions without making significant assumptions about the application domain or the dataset.

%% file: ethicalimpact.tex
Deep learning based decision making has ethical implications, especially in safety-critical applications, where failures could possibly lead to fatalities. This especially amplifies in shared settings like our application, where an agent’s decisions influence other agents’ decisions and so on. Certain features of our model contribute towards ethical decision-making. To begin with, our model is motivated by the need for autonomous agents to practice safety while navigating in human-centric environments. Our proposed framework takes into account the spatial influence of neighbors and implicit social navigation norms such as collision avoiding behavior that pedestrians follow when navigating in crowded environments towards predicting their future behavior. Further, our proposed framework acknowledges the multimodality of human motion and is capable of predicting multiple socially plausible trajectories per pedestrian in the scene. An autonomous agent that may use this framework to inform its navigation decisions would essentially take in to account all these multiple trajectories to negotiate a safe, collision-free path for itself. Often deep learning based models are reflective of inherent biases on the datasets that they are trained on. For instance, in our application, a model trained only on the UNIV dataset may not generalize well to a lower crowd density. However, as is the case with other baselines in our application domain, this is taken care of by using a leave-one-out approach, by training the model on four of five datasets and testing on the fifth. These datasets vary in crowd densities and contain a variety of trajectories of pedestrians interacting in several social situations, hence the training dataset is diverse. Moreover, a predicted trajectory can be mapped to the neighborhood (the learned domain) and hence, the neighbors that influenced the model's decision, hence providing some degree of interpretability to our framework. 

However, like all other deep learning models, our proposed framework relies on implicit assumptions that may have ethical consequences. For instance, our model relies on the assumption that the training dataset is reflective of ideal pedestrian behavior in shared environments or general pedestrian dynamics. Further, when deployed in a real-world setting to aid the navigation of an autonomous agent in a human centric environment, our framework's ability to predict intent accurately is largely dependent on the accuracy of input, i.e, the observed trajectory. Our model, by itself, does not account for the presence of adversaries that may provide deceptive input and cause our model to mispredict and cause undesired behavior. Further, in a real world setting, our model is expected to inform safety-critical decision-making of an autonomous agent in human-centric environments. Because deep learning models are black-box in nature, it is difficult to be able to completely ensure safety before deployment. It is therefore also important to incorporate a certain measure of confidence in the model's decisions, based on which its predictions can be followed or overridden. 